# The Rhetoric of Machine Learning[1]

Robert C. Williamson[2]

University of Tübingen

& Tübingen AI Center

**Abstract**: I examine the technology of machine learning from the perspective of rhetoric, which is simply the art of persuasion. Rather than being an neutral and "objective" way to build "world models" from data, machine learning is (I argue) inherently rhetorical. I explore some of its rhetorical features, and examine one pervasive business model where machine learning is widely used, "manipulation as a service."

Thirty-eight years ago, Clark Glymour claimed "artificial intelligence is philosophy."[3]

I agree. I further claim it is rhetoric[4]. It seeks to *persuade*, although it presents itself as just "*reporting the facts.*"

## FOUNDATIONS OF MACHINE LEARNING SYSTEMS

I'm an engineer / scientist / mathematician who works on theoretical questions arising in machine learning. I am also interested in how to think about socio-technical systems[5].

---

[1] Talk presented at AlphaPersuade 2.0, March 26, 2026, Irvine California. It is an extended version of a talk I gave at the 2nd Workshop on Learning Under Weakly Structured Information, 7 April 2025, with an earlier version presented on 12 November 2024 at Persuasive Algorithms? A symposium on the rhetoric of generative AI, in Tübingen. Some of the material was incorporated into the commentary "Environmental Intelligence: Context and Rhetoric" published in *Harvard Data Science Review* in 2025. Much of the material was originally developed for the Master's course I offered in Tübingen — *Beyond Fairness: A Socio-technical View of Machine Learning*, a complete recording of which is available on the Tübingen ML YouTube channel.

Since this document is a text of a talk design to stimulate, where the audience will be able to ask questions or disagree, it is deliberately written in a less hedged and careful style than a normal academic paper. I have deliberately pointed to a diverse literature, with the goal of challenging, inspiring, and opening new lines of inquiry, rather than merely justifying the same old stories.

[2] This work was funded by the Deutsche Forschungsgemeinschaft (DFG, German Research Foundation) under Germany's Excellence Strategy — EXC number 2064/1 — Project number 390727645, and by the BNBF through the Tübingen AI Center. Thanks to Kylie Catchpole and Laura Iacovissi for useful discussions.

[3] Clark Glymour, Artificial Intelligence is Philosophy, in James H. Fetzer (Ed.) *Aspects of Artificial Intelligence*, Springer, 195-207, 1988. More precisely he said "Artificial intelligence is philosophical explication turned into computer programs."

[4] For an alternate take on the rhetoric of "big data," see B. Mehlenbacher and A.R Mehlenbacher, The Rhetoric of Big Data: Collecting, Interpreting, and Representing on the Age of Datafication, *Poroi*, 16(1), 1-33, 2021. My motivation comes in large part from works on the rhetoric of science, such as Bruno Latour and Paolo Fabbri, "The Rhetoric of Science: Authority and Duty in an Article from the Exact Sciences," *Actes de la recherche en sciences sociales* 13, 81–95, 1977; translated to English in *Technostyle* 16(1), Winter 2000; James W. McAllister, Rhetoric of Effortlessness in Science, *Perspectives on Science* 24(2), 145–166, 2016; John A. Schuster and Richard R. Yeo, *The Politics and Rhetoric of Scientific Method: Historical Studies*, D. Reidel, 1986.

[5] Dating from my work on the report *Technology and Australia's Future,* Australian Council of Learned Academies, 2015.

I do not *build* AI systems, but I *study* them. I do not write *code*, but rather *proofs*.

I am *not* a rhetorician. However, I have found that looking at machine learning through the lens of rhetoric is refreshing and generative.

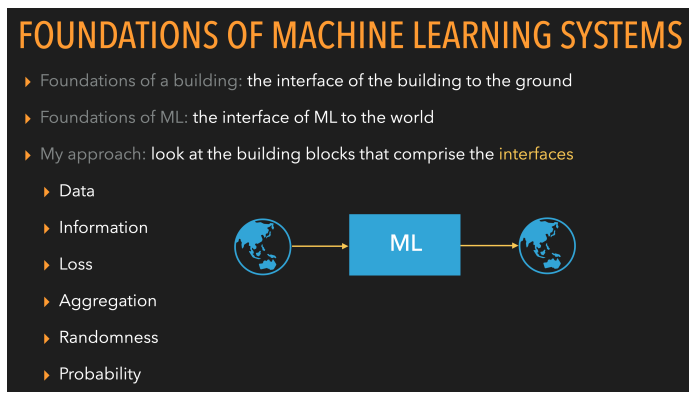

My starting point is the name of my chair: "Foundations of machine learning systems".

Rather than being something "perfectly solid" on which you can rely absolutely, I take "foundations" in the sense of the civil engineer's definition: *an interface between a building and the ground*. I thus look at the interface between ML and the world in terms of its mathematical primitives.[6]

My goal is to stimulate discussion along some different lines[7].

## ML IS RHETORIC

First up: by "rhetoric" I do not mean to examine the trite (and awful) imagery surrounding ML systems as depicted in the corny (machine-extruded) image presented here, which you can find countless variants of littering the internet.

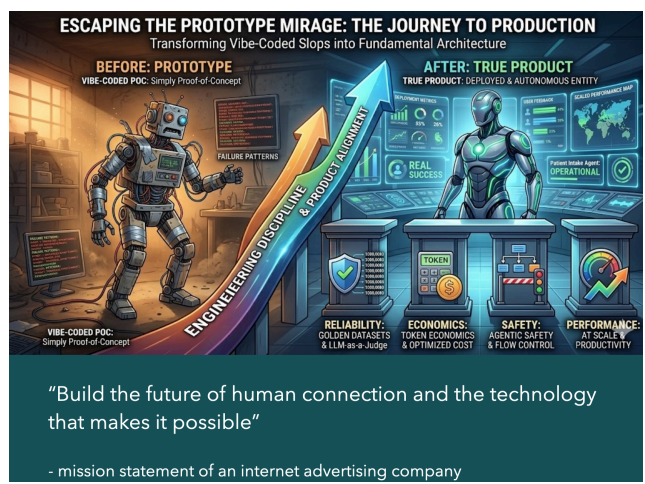

Nor will I consider the clearly rhetorical verbal statements about ML systems made by those who deploy them at scale for profit.[8]

In the same way that science[9], economics[10], and statistics[11] have all been fruitfully viewed from the perspective of rhetoric,[12] it is instructive to examine the "rhetoric of the technology of machine learning."

---

[6] The basic building blocks of these interfaces are data, information, loss, aggregation, randomness and probability.

[7] Rather than telling you the "one true way" of thinking about machine learning, because, of course, there is no such thing! Nor do I imagine I have achieved any sense of completeness; I'm happy if I have initiated new conversations.

[8] The droll example illustrated is the mission statement of Meta. That Meta is an advertising company is clear from their 2025 Q1 earnings release (page 8): 97.8% of their revenue comes from advertising.

[9] Walter B. Weimer, "Science as a rhetorical transaction: Toward a nonjustificational conception of rhetoric." *Philosophy & Rhetoric* 10.1 1-29, 1977; R. Allen Harris, Rhetoric of Science, *College English* 53(3), 282—307, 1991; Henry Krips, J.E. McGuire and Trevor Melia (Eds), *Science, Reason, and Rhetoric*, University of Pittsburgh 1995.

[10] Deirdre N. McCloskey, *The Rhetoric of Economics* (2nd edition), The University of Wisconsin Press, 1998.

Machine Learning is indeed a *technology*.

But unlike many technologies it is a *cognitive* technology[13].

It thinks *on our behalf.* Not on its *own*. We *delegate* certain tasks to it, and then let it persuade us of its results. We offer it data, and then pay attention to what it "generates" (predictions or forecasts or explanations).

## STYLES OF REASONING

Or as Ian Hacking put it, "the style of reasoning" of ML. Hacking identified several different styles of reasoning used in science[14]. They are incommensurate, perhaps contradictory, but all are widely used and underpin the scientific advances we are lucky to have.

He looked closely at the "statistical style,"[15] which ML shares much of — the creation of new classes or categories, new law-like claims, new objects (the distribution, and its properties - means, variances etc), new criteria for reliable reason, and circularity ("assertions about probability are themselves assessed using probabilities"), and new technologies of objectivity[16].

---

[11] Deidre McCloskey,. "Rhetoric within the citadel: Statistics." In *Argument and Critical Practice: Proceedings of the Fifth SCA/AFA Conference on Argumentation*, pp. 485-490. 1987.

[12] By which I simply mean argument intended to persuade. We (or least some seem to) let ML persuade us by accepting its outputs on its own terms, rather than demanding a richer and more robust form of persuasion.

[13] Just as written language is. See Eric A. Havelock. *The Muse Learns to Write: Reflections on Orality and Literacy from Antiquity to the Present.* Yale University Press, 1986; Eric A. Havelock, *The literate revolution in Greece and its cultural consequences.* Princeton University Press 1982; Jack Goody, Technologies of the intellect: Writing and the written word. In *The Power of the Written Tradition*, pages 132–151, Smithsonian Institution Press, 2000. See also the references to Walter Ong further below.

[14] Namely mathematical, experimental, hypothetical modelling, taxonomic, statistical, historical-genetic, and laboratory sciences.

[15] Ian Hacking, Statistical Language, Statistical Truth and Statistical Reason: The Self-Authentication of a Style of Scientific Reasoning, pages 130—157 in *Social Dimensions of Science*, University of Notre Dame Press, 1992.

[16] An outstanding analysis of the "actuarial stance" that is used in statistics, and less reflectively in machine learning, is the wonderful book by Alain Desrosières, *The Politics of Large Numbers: A History of Statistical Reasoning,* Harvard University Press, 1998.

Viewing ML from the perspective of insurance is surprisingly generative; see Christian Fröhlich and Robert C. Williamson. "Insights from insurance for fair machine learning." In *Proceedings of the 2024 ACM Conference on Fairness, Accountability, and Transparency*, 407-421, 2024.

A crucial aspect of the statistical style, that is inherited by ML, is what I call "the *actuarial stance*"[17].

Imagine a town with many clubs. Each club has many members. Each person belongs to many clubs. It is natural for us to identify a given club with its set of members. The actuarial stance takes the dual[18] view of identifying each individual *with the set of clubs they are members of*.

That is ML's view of you: *the set of clubs of which you are a member.* Nothing more.

## THE STYLE OF REASONING OF MACHINE LEARNING

A cartoon version of the style is given on the slide. I want to concentrate on two aspects.

> ▸ ML just extracts information from data
> ▸ It consequently uncovers the "intrinsic structure" of the data
> ▸ Data is given, and the ML user's job is to just process it
> ▸ Data is "drawn" from a probability distribution
> ▸ Data is a stable and solid representation of the world: a fact
> ▸ The ML algorithm thus learns an objective representation of the world

Like the law, and science for that matter, the rhetorical style of ML is primarily that of "*anti-rhetoric*[19]," which is to say it explicitly denies that it is rhetoric. It is not (so the consensus would insist) trying to *persuade* anyone, but rather it is a "view from nowhere,"[20] just revealing what the data showed on entirely on its own — the "*intrinsic structure of the data*".[21]

But data is *always* the end result of a chain of operations, and without knowing them, all you have is a string of bits with no "intrinsic" structure, and from which it is impossible to make a persuasive argument.

---

[17] Because this is exactly the view of insurance companies, where actuaries work. The "clubs" need not be real clubs. For example, they can be people 35-37 years old, or those having a particular gene variant, or those who are known to like cats. But the stance remains the same: from ML's perspective, you are nothing but the set of clubs you are a member of.

Thus your identity (from ML's perspective) depends crucially on which clubs have known membership lists! See also the discussion below on Counting and Categories.

[18] The setup can be mathematically formalised as a hypergraph (a set system) and the "dual" I mention is literally the dual hypergraph.

[19] Gerald B. Wetlaufer, Rhetoric and Its Denial in Legal Discourse, *Virginia Law Review,* Vol. 76, No. 8, 1545-1597, Nov. 1990.

[20] The phrase "view from nowhere" is due to Thomas Nagel; the idea is that rather than having a *viewpoint* (where you stand, which implies "biases") the ML system is "objective" (i.e. *viewpoint-free*). Of course this is nonsense. The choice of what data to take as given is already a substantial and significant viewpoint choice.

[21] But there is *no such thing* as the "intrinsic structure of the data," notwithstanding the commonality of this phrase! Try an internet search for the quoted phrase to see its pervasiveness.

ML is described as being "data-driven" and this is held to be a very fine thing as it (supposedly) avoids the biases that people have. It is "objective" and reveals the world as it really is. Well so it is claimed…

## SELF-AUTHENTICATION AND ENTHYMEMES

The power of this stance is that it is (to use Hacking's phrase) "self-authenticating"; styles of reasoning, once in place, disable us from thinking differently:

> The truth is what we find out in such and such a way. We recognise it as truth because of how we find it out. And how do we know that the method is good? Because it gets at the truth. …
>
> A style of reasoning, once in place, is not relative to anything. It does not determine the standard of objective truth. It is the standard.[22]

The scientific paper has the rhetorical structure of an *enthymeme*[23], whereby a major premise is left entirely implicit, namely the scientific "background knowledge".

So too is the rhetorical structure of ML an enthymeme, or if not, a major premise is at least swept under the carpet and not examined. I will now examine several enthymemes of ML.

## DATA AS FACT

All arguments start somewhere. And that place is the *fact*.[24]

The "fact" underpinned the invention of statistics, on which machine learning builds.

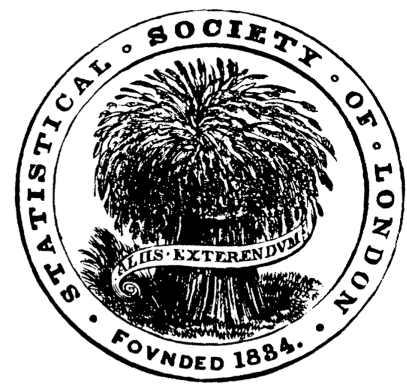

The prospectus of the Statistical Society of London (1838) stated

> The Statistical Society will consider it the first and most essential
> rule of its conduct to exclude carefully opinions from transactions and publications.[25]

---

[22] Hacking,

[23] Z. Livnat, The Concept of Scientific Fact: Perelman and Beyond, *Argumentation* 23, 375-386, 2009.

[24] Interestingly, the idea of "the fact" had to be invented, and its history has been well described by Mary Poovey, *A History of the Modern Fact: Problems of Knowledge in the Sciences of Wealth and Society,* The University of Chicago Press, 1998*;* see also Barbara J. Shapiro, *A Culture of Fact: England, 1550—1720,* Cornell University Press, 2000.

[25] Quoted on page 47 of "The Exclusion of Opinions," *The London and Westminster Review*, April-August 1838.

Their motto was *aliis exterendum* — "to be threshed out by others."[26]

The logic was that their data (statistics) were "just the facts", and once collected, could be reliably built upon (by others).

## REFRAMING FACTS

Rather than construing the fact as *given* (after all, its etymology comes from the latin word "to *make*"!), students of rhetoric (in particular Chaim Perelman[27]) define the fact as the *beginning* of the argument and, crucially, *that which one <u>chooses</u> not to question.*[28]

For ML, *data is its fact*[29].

It is "given" — the Latin root of data is *dare*, to give. And it is (largely) taken for granted[30].

An *argument* is a network of claims underpinned by a warrant (which certifies that the argument is adequately sound and valid)[31]. The traditional view of ML only pays attention to the final step of the argument.[32]

---

[26] Victor L. Hilts, *Aliis exterendum,* or, the Origins of the Statistical Society of London, *Isis*, Vol. 69, No. 1, pp. 21-43, Mar., 1978; the translation by Hilts (which I quote) has been challenged. Some have offered "for others to thresh" or "for others to thresh out" or "let others thrash it out". Vic Barnett, (Aliis Exterendum and Beyond! *Radical Statistics* 98, 68-69, 1998) offers the quite different meaning variant: "to be threshed *for* others". This shows the danger of adopting Latin mottos!

But we should not be distracted by these linguistic quibbles, as the detailed history of the society (recounted by Hilts and others) makes it clear that their view was that they just wanted to collect the data, and that they viewed this as intrinsically worthwhile because it was the collection of *facts*.

[27] Chaïm Perelman and Lucie Olbrechts-Tyteca. *The New Rhetoric – A Treatise on Argumentation.* University of Notre Dame Press, 1969.

[28] Ludwik Fleck, in his Genesis and Development of a Scientific Fact nicely defined it as "a stylised signal of resistance in thinking", which gets to the bottom of the matter. See Ludwik Fleck, *Genesis and Development of a Scientific Fact,* The University of Chicago Press, 1979.

[29] Confer Daniel Rosenberg, Data Before the Fact, in *Raw Data is an Oxymoron,* pages 15-40, MIT Press 2013 and Jonathan Furner, "Data": The Data, pages 287-306, in *Information cultures in the digital age: A festschrift in honor of Rafael Capurro.* Wiesbaden: Springer Fachmedien Wiesbaden, 2016.

[30] This is nicely demonstrated in Nithya Sambasivan, Shivani Kapania, Hannah Highfill, Diana Akrong, Praveen Paritosh, and Lora M. Aroyo. "'Everyone wants to do the model work, not the data work': Data Cascades in High-Stakes AI." In *Proceedings of the 2021 CHI Conference on Human Factors in Computing Systems*, 1-15, 2021. Many LLMs do not declare the data they were trained upon (or like Llama, just refer to the "<u>pile</u>", an unspecified collection of pirated texts.)

[31] Stephen Toulmin, *The Uses of Argument*, Cambridge University Press, 2003.

[32] That science requires the whole chain (or better said, network or tangle) is nicely argued in Nancy Cartwright, Jeremy Hardie, Eleonora Montuschi, Matthew Soleiman and Ann CV. Thresher, *The Tangle of Science: Reliability Beyond Method, Rigour, and Objectivity*, Oxford University Press, 2022.

Rather than taking facts (or data) as *given*, we can demand their "*chains of reference*" — the string of claims which provide a warrant for the final assertion. Such chains, or networks, are a stand-out distinguishing feature of scientific knowledge.[33]

## MODELS AS REPRESENTATIONS

Machine learning purports to "represent the world": Ilya Sutskever, chief scientist at OpenAI, said

> When we train a large neural network to accurately predict the next word in lots of different texts…it is learning a world model…. This text is actually a projection of the world…. What the neural network is learning is more and more aspects of the world, of people, of the human conditions, their hopes, dreams, and motivations…the neural network learns a compressed, abstract, usable representation of that.[34]

But representations can not be well crafted by mere observation, as science has well discovered. One needs to give the world a *kick*[35]; to "intervene."[36]

The rhetoric of ML holds rather that representations of the world can be constructed by mere observation of convenience samples — so much so that one of the main ML conferences is called "Conference on Learning Representations."[37]

The trouble with viewing models as "representations" is that the word suggests too strongly that they are true. We should instead think of them as (useful) fictions.[38]

---

[33] Bruno Latour, "Circulating Reference: Sampling the Soil in the Amazon Forest." pages 24–79 in *Pandora's Hope: Essays on the Reality of Science Studies*, Harvard University Press, 1999.

[34] Quoted in Alex Reisner, AI's Memorization Crisis, *The Atlantic*, 9 January 2026.

[35] It is not a matter of merely "projecting". Of course, representations are pervasive and immensely valuable in science; we can hardly understand the world without them. My complaint is not about representations, but at the glibness of the view that ML models, especially ones that were just trained on *words,* have made representations that one can rely upon to interact with the world (such is the rhetoric of the idea of "foundation models"). For the complexities of scientific representation, see Michael Lynch and Steve Woolgar (Eds), *Representation in Scientific Practice*, MIT press, 1990; and Catelijne Coopmans, Janet Vertesi, Michael Lynch, and Steve Woolgar (Eds), *Representation in Scientific Practice Revisited*, MIT Press, 2014.

[36] in Hacking's phrase. And in any case, "Theories, not individual sentences, are representations." (Page 134 of Ian Hacking, *Representing and Intervening: Introductory Topics in the Philosophy of Natural Science,* Cambridge University Press 1983). The problems with the view that representations are readily found and accurate depictions of the world is nicely summarised in chapter 2 of Steve Woolgar, *Science: The Very Idea*, Ellis Horwood, 1988.

[37] There is also much hype around "foundation models" on which one can build all manner of inferences. I will return to this point later in my discussion of (reversible) black boxes.

[38] Thus following the philosophy of modelling that can trace its lineage to Hans Vaihinger's *The Philosophy of 'As If'*, Routledge, 1924.

## THE COMFORT OF RANDOMNESS

Another pervasive enthymeme is that data is "drawn independently from some probability distribution."[39]

This is important for much ML because its methods are built on the mathematics of classical probability theory[40]. Indeed, *if* one's data is so "drawn", *then* there are many mathematical results one can rely upon, such as the misleadingly named "law" of large numbers[41].

The problem is that much of the data used by ML is most certainly not so "drawn." Rather it is a "convenience" sample of whatever could be found lying around. (This matters a lot: Facebook's vaccine hesitancy study had a sample size of 250,000, but this was as informative as an actual "random sample" of size ten[42].)

Why is this strange assumption so popular? Because via the simple incantation of "drawn iid" *one need not worry about the long chain that created your data* — you can get on with playing with your models — and models are sexy, right?

Why does this matter?

Because it greatly affects the results! If one tries to more deeply understand this "drawn independently" notion one is led to try and understand *randomness*.

Widely used as an undefined primitive that is presumed to nevertheless have a clear meaning, it *is* actually amenable to analysis. When one does so, one finds there is no single universal "randomness" — such a notion is logically void. What we find instead is an infinite family of different types of randomness; which one you adopt is a choice.

---

[39] For a take on such assumptions, see Richard A. Berk and David A. Freedman. Statistical assumptions as empirical commitments. pages 235-254 in Thomas G. Blomberg and Stanley Cohen (Editors), *Law, punishment, and social control: Essays in honor of Sheldon Messinger* (Enlarged Second Edition), Aldine de Gruyter, 2003.

[40] Ironically, the father of the mathematical theory of probability that is widely appealed to in ML, Andrei Kolmogorov, was acutely aware that there was no reason to presume data in the world has a "probability" associated with it; it is an *assumption* not a fact. See Andrei N. Kolmogorov, "On Logical Foundations of Probability Theory," pages 1-5 in Jurii V. Prokhorov and Kiyosi Ito (Editors), *Probability Theory and Mathematical Statistics: Proceedings of the Fourth USSR–Japan Symposium,* 1982.

[41] Misleading because it is not an empirical "*law of nature*" (to use the now outmoded phrase), but simply a *mathematical theorem* which can not make any legitimate claims about the world. See the discussion of this point in Appendix D of Christian Fröhlich, Rabanus Derr, and Robert C. Williamson. Strictly frequentist imprecise probability. *International Journal of Approximate Reasoning* 168 (109148), 2024.

[42] Valerie C. Bradley, Shiro Kuriwaki, Michael Isakov, Dino Sejdinovic, Xiao-Li Meng, and Seth Flaxman. Unrepresentative big surveys significantly overestimated US vaccine uptake. *Nature* 600(7890), 695-700, 2021.

This choice matters, and it matters *ethically*.

It turns out[43] that randomness and fairness are essentially the *same thing*: if you accept that the choice of a "protected attribute" (e.g. race or sex) is a consequential choice, then you are logically bound to draw the same conclusion regarding randomness, because the very definition of fairness in ML reduces to a particular kind of independence, which is a type of randomness.

And this "independence" is on shaky ground: it is used to justify the move from individual samples to a distribution, but it is defined in terms of the *very thing that it is used to justify — the distribution.*[44]

I claim, to the contrary, that there is no good reason to presume that ML's data is "generated by sampling from some probability distribution"[45].

## COUNTING AND CATEGORIES

ML is built on statistics, and statistics works with aggregates. We count.[46]

But *what* do we count? We count according to predefined categories — this many white people, this many black. Then we can compare our counts. That's statistics.

---

[43] Rabanus Derr and Robert C. Williamson, Fairness and Randomness in Machine Learning: Statistical Independence and Relativization, *The New England Journal of Statistics in Data Science* 3(1), 55-72, 2025.

[44] This circular self-justification is but one instance of that move as we shall see below.

These subtleties cause much grief even for very simple prediction problems (who is best suited to get a college placement offer?). What they mean for the more complex uses of LLMs is hard to imagine.

More sophisticated notions of randomness have been defined, e.g. due to von Mises and Vovk, more reasonably, in terms of the data itself, rather than the theoretical entity of the distribution. For pointers to the literature on randomness, see Derr and Williamson (op. cit.).

[45] This is perhaps the most challenging claim I make (for the majority of ML scientists) based on my experience so far. It is usually met with incredulity: "what do you *mean* the distribution does not exist?!" The simple truth of the matter is that there is no reason for it to exist, other than for the convenience of the scientist. Its existence is especially fraught when the data is about people. But I am not alone in challenging its existence; see Ben Recht, *There is no Data-Generating Distribution*, https://www.argmin.net/p/there-is-no-data-generating-distribution (December 2025) and Benedikt Höltgen and Robert C. Williamson, *The costs of pretending that there are data-generating distributions in the social world,* to appear in FaCCT 2026.

[46] Recall my depiction of the "actuarial stance" earlier. *Exactly* how the counting is done, and more specifically, *how counts of different categories might be combinable*, makes visible another pervasive enthymeme in ML (and more generally in statistics): it is usually taken for granted, that the set of things one can count ("measure") form an *algebra* — a systems of sets closed under complementation, and arbitrary intersections and unions. *This is often not justified.* If one relaxes the usual strong assumption, one ends up with a richer mathematical theory which leads unavoidable to *imprecise* probabilities for many of the events — this is tantamount to the fact that the counting is *necessarily imprecise.* (See Rabanus Derr and Robert C. Williamson, Systems of Precision: Coherent Probabilities on Pre-Dynkin Systems and Coherent Previsions on Linear Subspaces, *Entropy* 25, 1283, 2023.)

In one of the earliest books on statistics (*The Logic of Chance*[47]), John Venn so recognised the importance of the choice of categories (you can't accumulate counts if you do not what you are counting) that he coined the notion of a "natural kind," as a name for the idea that categories are actually out there in the world[48]. ML has invested heavily into this notion!

## THE ONE AND THE MANY

ML algorithms work with aggregates; but they typically have their consequences on individuals, who can be placed into many *different* aggregates[49]. To be sure, sometimes one cares about the *aggregate* itself, but in the most emotionally and politically charged cases, it is *individuals* that matter[50].

The connection between the aggregate and the individual is largely ignored in almost all uses of ML and it is taken for granted that reasoning at the aggregate level suffices.

Some statisticians[51] are aware of the problem, under their name for it "the ecological fallacy[52]" — the fallacy being reasoning from the group to the individual: what can you conclude about *me* if the conference organisers tell you that 50% of the attendees are vegetarian?

*Nothing.*

Interestingly, this is simply the dual to the problem of randomness (which assumes away the difficulties of going from the individual to the aggregate).

---

[47] John Venn. *The Logic of Chance, an essay on the theory of probability*. MacMillan and Co., London, 1876.

[48] Rather than being a construct of our minds; see the fascinating book George Lakoff, *Women, fire, and dangerous things: What categories reveal about the mind.* University of Chicago press, 2008.

For a more detailed summary of the implications of categories for machine learning, especially from the ethical perspective, see Appendix H of Aditya Krishna Menon and Robert C. Williamson. "The cost of fairness in binary classification." In *Conference on Fairness, Accountability and Transparency*, pp. 107-118. PMLR, 2018; the appendices, which are frustratingly separated from the main paper, are available here.

[49] This is often called "the reference class problem" as if it is a problem that can be solved. See Alan Hájek, "The reference class problem is your problem too." *Synthese* 156, no. 3, 563-585, 2007.

[50] Perhaps old style soviet social planners could exclusively concern themselves with the welfare of the aggregate (or at least *say* that is what they were doing). But essentially all ethics is about individuals.

[51] And some psychologists: E. D. Beck, & J. J. Jackson, Idiographic Traits: A Return to Allportian Approaches to Personality. *Current Directions in Psychological Science*, 29(3), 301-308, 2020.

[52] https://en.wikipedia.org/wiki/Ecological_fallacy

But the single move done in most ML is to deal with this simply by use of the phrase "probability of" — this individual "has a probability of X" of some outcome, and all of the difficulties are buried in the undefined semantics of that single misunderstood word.

## INFORMATION , DATA AND LEARNING

Machine learning claims to extract "information from data", and it does so by "learning". Surely there is nothing rhetorical here?

The trouble starts with with "information." A formal definition was developed by Claude Shannon[53] *specifically* to solve the problem of reliable communication. One aspect of this is "source coding," whereby messages are "compressed" to use less bandwidth. We are all familiar with this through jpeg images and gzipped files.

Think of the "information" preserved when one goes from manuscript, to physical printed book, to the text itself, to its octal representation, to its gzipped form. *Something* is persevered, and that something is *information*.

What is the connection with machine "learning"?

Well, "language modelling *is* compression"[54]. This should be hardly a surprise: the basic quantity that is optimised in training large language models is "perplexity", which is a transformed version of Shannon's *entropy,* which in turn controls the degree to which a data source can be compressed — it captures the intuition that simple patterns can be compressed a lot by simply describing the pattern succinctly.

The irony is, notwithstanding the way the training algorithms work[55], businesses built upon machine learning *explicitly deny* they are processing and compressing information.

OpenAI claims: "models do not store copies of the information that they learn from."[56]

---

[53] Claude E. Shannon, "A mathematical theory of communication." *The Bell System Technical Journal* 27, no. 3, pages 379-423 and 623—656, 1948. There are many variants (which itself is a significant point), but they all share a "family resemblance." See Yury Polyanskiy and Yihong Wu, *Information Theory: From Coding to Learning,* Cambridge University Press 2025.

[54] Grégoire Delétang, et al. "Language modeling is compression." *arXiv preprint arXiv:2309.10668*, 2023.

[55] Part of the irony is that the other framing of what ML is doing (minimising the expected loss of a predictor) turns out to be exactly equivalent to optimising a notion of information. More precisely, information and expected risk are in a one-to-one relationship. See Mark Reid and Robert Williamson. "Information, divergence and risk for binary experiments." *Journal of Machine Learning Research* 12, 731—817, 2011; Robert C. Williamson and Zac Cranko. "Information processing equalities and the information–risk bridge." *Journal of Machine Learning Research* 25, no. 103, 1-53, 2024.

[56] From Reisner, op. cit.

This is quite funny given that Sutskever (their chief scientist) said (as I quoted before) "the neural network *learns a compressed, abstract, usable representation* of [the world]."

Similarly, Google claims: "there is no copy of the training data—whether text, images, or other formats—present in the model itself."[57]

These claims fall flat when you realise it is possible to largely reconstruct *entire books* by repeated queries to large language models.[58] So we could say that they have not "*learned*" anything, so much as simply *compressing the information.*[59]

## AN OBSESSION WITH METHOD

Beyond the above enthymemes, there is an even deeper rhetorical presumption of ML, and that is its *obsession with method.*

The logical positivists fantasised that if they got their protocol sentences and the like in order, they could develop an "unbiased" method of revealing the truth about the world. They have largely disappeared from philosophy … because they now work for Machine Learning companies. But their love of method lives on.

Recalling again Sutskever's claim quoted earlier of "learning a model of the world" by "predicting the next word in lots of different texts". Such a stance has been well criticised:

> Something in the relationship between science and reality has been misconstrued. Literary expression is allowed to masquerade as physical reality. And in the process, literature and language have been utterly misrepresented. Literature has been atomized and expression reduced to a mere reassembling of interchangeable parts.[60]

---

[57] Ibid. There is indeed no complete exact copy, but they don't need that, in the same way as one does not need to photographically reproduce a book to infringe copyright; it suffices to reproduce the words, and even if there are changes of spelling or punctuation, or even the omission of entire chapters, nobody is fooled as to what is going on.

[58] Ahmed Ahmed, A. Feder Cooper, Sanmi Koyejo, and Percy Liang. "Extracting books from production language models." *arXiv preprint arXiv:2601.02671,* 2026.

[59] Notwithstanding the claims made by those who profit from the exercise.

The compression is lossy (as Delétang et al., op. cit., elaborate); the books can not be extracted perfectly, but well enough to make it clear that large parts of the information in the training data has been effectively stored in the model.

[60] The quote is from Walter J. Ong, Ramist classroom procedure and the nature of reality. *Studies in English Literature, 1500-1900* 1(1), 31-47, 1961.

This was written by Walter Ong … *in 1961!* [61]

How was he so prescient? Because he was criticising Ramism, the doctrines of 16th century "professor of eloquence" who is the grandfather of all obsessions with method, and thus the patron saint of machine learning in its current guise.[62]

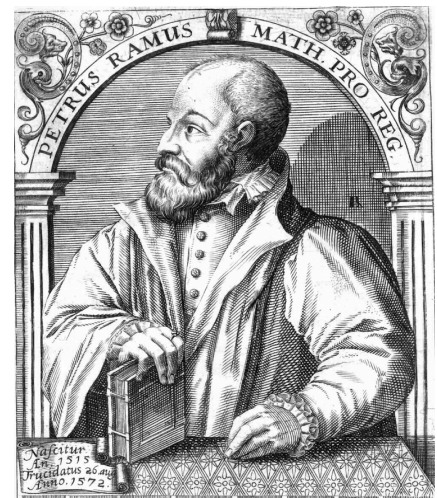

Ramus, made no great discovery nor exhibited any great originality. The thing he *did* do was to *valorise method*, and overly simplistic dichotomies.[63]

For our purposes today, what matters is that, in Ong's words (which well describe LLM practices)

> Ramist 'method' makes it possible to think of knowledge itself in terms of 'intake' and 'output' and 'consumption' …[64]

The overarching Ramist (and ML) goal is one of *efficiency*: "the mysterious realm of knowledge was reduced to something one could manage, almost palpably handle. The arts and sciences could be viewed as a mass of '*wares*'."[65]

Ramus can well be blamed for the "decline of agonistic practice"[66] — like bad teachers the world over, he did not like disagreement or debate, but wanted his students and acolytes to just learn the way the world was *according to his daft diagrams.* Ramus considered his method as "uncontaminated from rhetoric"[67] and held that knowledge was essentially a set of facts, that can be captured by one of his convoluted dichotomies.

---

[61] Ong seems apposite for our current endeavour — the subtitle of his arguably most famous book is *Technologizing of the Word.:* Walter J. Ong, *Orality and Literacy: The Technologizing of the Word,* Routledge 2002. See also his "Writing is a technology that restructures thought," pages 293—319 in *The Linguistics of Literacy,* John Benjamins Publishing Company, 1992.

[62] The full story of Ramus is the subject of the delightful book by Walter J. Ong, *Ramus, method, and the decay of dialogue: From the art of discourse to the art of reason*. University of Chicago Press, 2004. For an alternate take (which I do not find fully convincing) see Philippe Hamou. Sur les origines du concept de méthode à l'âge classique: La Ramée, Bacon et Descartes. *Revue LISA / LISA e-journal*, XII (5), 2014 available at https://hal.parisnanterre.fr/hal-02326900v1

[63] Chapter 9 of Ong's book gets to the core of the matter: "the method of method." He observes that "Modern interest in scientific method has its matrix in this mélée" (page 228).

[64] Walter J. Ong, Ramist method and the commercial mind. *Studies in the Renaissance* 8, 155-172, 1961, (page 161).

[65] Ibid, page 169. Ramus is largely responsible for the spatialisation of knowledge, a virulent meme that we are still immersed in today:
> "all these approaches to knowledge are approaches congenial to persons who habitually deal with reality in terms of accounting rather than in terms of meditation or wisdom. 'Method' is an early step in the procedures which encode knowledge in a neutral, leveling format, reducing it to bits of information such as those which will eventually make their way into electronic computers." Ibid, pages 171—172.

[66] This lovely phrase is due to Melinda Farrington, Ong's Ramus: Origins and Implications of the Decline of Agonistic Practice, *Western Journal of Communication,* 86:5, 716-733, 2022. Her point, very apposite for the understanding of the effect of ML systems, is that the Ramist stance rejects the very notion of agonism — debate, disagreement, discussion (conversation in a word), in favour of an overly authoritative declaration of the "truth" in the form of a confident answer.

The full history is complex[68], but suffice to say, the Ramist method underpins the *commodification*[69] of knowledge, the latest incarnation of which are LLMs.[70]

Nowadays, machine learning valorises method ("algorithms"[71]) even more than Ramus[72], and these algorithms are judged, ironically enough, according to yet more rigid methods![73]

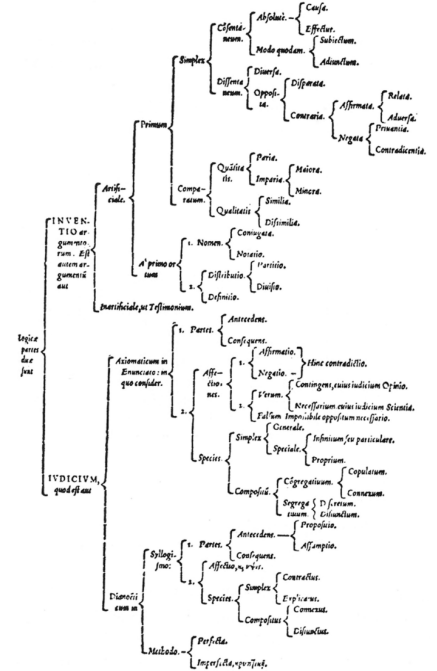

## BENCHMARK DATA SETS — NAIVE EMPIRICIST AND PROUD OF IT

A pervasive feature of machine learning practice is the "benchmark data set" and the associated numerical indices of performance attained by algorithms on it[74].

---

[67] Farrington, op. cit. page 721

[68] Others have started to grapple with the pre-history of the modern valorisation of method. See for example Justus Buchler, *The Concept of Method*, Columbia University press, 1961; Henri Stephanou, *Systems, Machines, and Problems Solving*, De Gruyter, 2025; Lorraine Daston, *Rules: A Short History of What We Live By*, Princeton University Press, 2022. Of course the development of the notion of method has been of immense value; the danger is that it becomes the only way we approach the world. There is also a developing literature that pushes back against method, most famously by Paul Feyerabend, *Against Method*, Verso, 1993; but also see John Law, *After Method: Mess in Social Science Research*, Routledge, 2004, and Gareth Morgan (Ed.), *Beyond Method: Strategies for Social Research*, Sage Publications, 1983 for attempts by social scientists to transcend the constraints of "method."

[69] At bottom, "The Ramist method 'thingifies' language and knowledge," [Ibid, page 730] and seems to me to be behind much of the rhetoric of ML.

[70] In particular, his dichotomies "left no room for divergent meanings of particular words or ideas, suggesting instead 'a one-for-one correspondence between terms and things'" Ibid, page 728; note the resonance with the earlier claims of representation.

[71] For a breathlessly enthusiastic (but perhaps now somewhat dated) example of the valorisation of algorithms in machine learning, see Pedro Domingos, *The Master Algorithm: How the Quest for the Ultimate Learning Machine Will Remake Our World*, Basic Books, 2015.

[72] Its conferences are filled with new methods and (relatively) little else.

[73] It is worth stressing: approaching the solution of problems in a *systematic* way (which is what method does — see Stephanou, op. cit) is immensely valuable. But when one gets too enamoured of one's methods, and thus ignores everything that does not fit neatly into a well-specified method, one runs the risk of fooling oneself badly.

[74] The graph on the slide is from Alexander Braun, Tim Schreier, Luis Winkelmann and Sebastian Zezulka, "Benchmark data and problem solving in machine learning research", course paper for *Beyond Fairness: A Sociotechnical Systems view of Machine Learning*, University of Tübingen, 2022. I do not have more recent data, but it is a safe bet that the curves kept climbing.

Pictured are two ML engineers benchmarking their latest algorithm … which has been optimised for the benchmark (and *nothing else*). All that matters is some scalar performance.

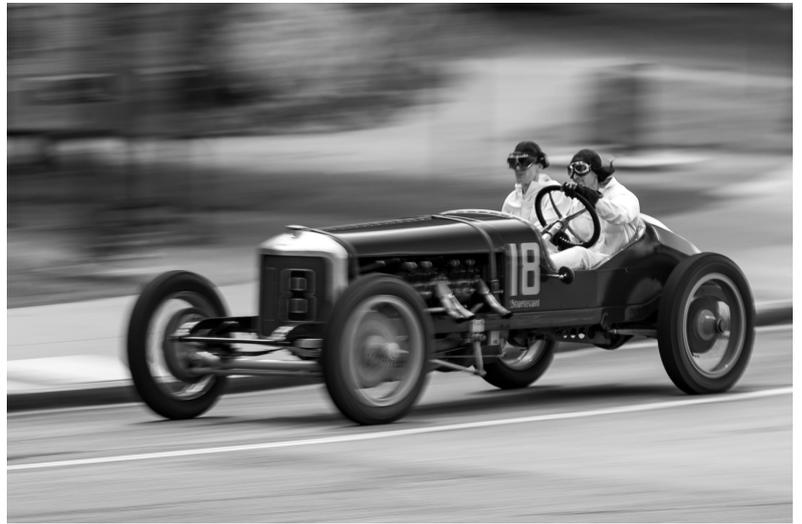

If your paper submitted to a ML conference does not improve on "SOTA" on a standard benchmark, then the referees will castigate you for your lack of adherence to the religious practices that have been ordained[75]. (SOTA is "state-of-the-art" and its widespread acronymisation reinforces my point)[76].

Funnily enough, there are plenty of folks writing apologetics[77] for benchmarks, which however, can be seen merely as a manifestation of the self-justification of a style of reasoning — the use of benchmarks is seen to lead to a success; the success is used to justify the benchmarks.[78]

---

[75] Calling this a religious practice might seem like hyperbole; but any scientist who is not infected with the mania for SOTA leaderboards and who has interacted with anonymous ML referees in the major ML conferences would consider it fair. If you prefer, you could call it an "observational ideology" (page 52 of Paul Feyerabend, *Against Method*, (3rd edition) Verso (1993).) He identified the problem long ago: "[S]ome of the methods of modern empiricism which are introduced in the spirit of anti-dogmatism and progress are bound to lead to the establishment of a dogmatic metaphysics and to the construction of defence mechanisms which make this metaphysics safe from refutation by experimental inquiry," from page 4 of Paul Feyerabend, "How to be a good empiricist: a plea for tolerance in matters epistemological," pages 3–39 in B. Baumrin (ed.), *Philosophy of Science: The Delaware Seminar, volume 2*, Interscience Press, 1963.

[76] One might speculate that a large part of their appeal is the inherent ordinalization they offer: "Quantification is a way of making decisions without seeming to decide," as is their appeal for the modern inept manager [Isabelle Bruno, "The "indefinite discipline" of competitiveness benchmarking as a neoliberal technology of government." *Minerva* 47, no. 3, 261-280, 2009.]

[77] I use the word as in https://en.wikipedia.org/wiki/Apologetics ; such apologetics can be seen as disciplinary "boundary maintenance" — the solidifying and protecting of nascent disciplinary boundaries.

[78] One recent example: is Ben Recht's Benchmarking Culture, 10 March 2026. The argument is made that obsession with benchmark performance (and nothing else) has led to the advances in ML — all that matters is performance. The similarity with J.B. Watson's arguments in favour of behaviourism in psychology are uncanny: Watson "was originally able to convince his colleagues that the discipline could exhibit cumulative growth if it conducted research whose results would be unaffected by whichever theory of introspection or neurophysiology turned out to be correct." [Steve Fuller, *Social Epistemology,* (2nd edition), Indiana University Press, 2002, page 200.]

This stance is not merely "naive empiricist and proud of it" (to use my earlier phrase), but offers denial against any critique, by *empirically* justifying how good the benchmarking culture is; Ian Hacking was dead-right — it is the *self-justification* of a style of reasoning. Within this bubble, there is no escape possible (and those inside see no reason to try to escape!). You need to step *outside* of the bubble, which is a move that most ML engineers do not consider as an option — after all, their stuff "works", right?

Although some recognise problems with the approach, the suggested mitigations are to *improve* benchmark practice, rather than to *transcend* them.[79]

I perceive several problems:.
- Benchmarks are presumed to be *representative* (there's that "factyness"again)
- Performance is (usually) judged in terms of predictive accuracy *alone;*
- Benchmarks often are made from "convenience samples" (data you find just lying around).

As well as the problem of *scalarisation*[80] (only caring how fast your car can go, not all the many other dimensions that matter), the "benchmark" obsession further cements the idea of a data "*set.*" A set is a thing. A pervasive presumption is that data is actually a *thing,* rather than a *process* which needs to be understood and documented.

Contrast the approach to data of much of ML with that of a careful empirical scientist who takes care in collecting the data. The ML user will often just download it. All that matters is the numbers. Questions of its provenance, reliability, choice of categorical labels are often ignored as inconsequential.

## ML AS BLACK BOX

So what is an alternative?

Consider both ML systems, and data, as a *process*[81], or to use a more powerful metaphor of Latour, as "*reversible black boxes*[82]".

---

[79] See for example Maria Eriksson, Erasmo Purificato, Arman Noroozian, Joao Vinagre, Guillaume Chaslot, Emilia Gomez, and David Fernandez-Llorca. "Can we trust AI benchmarks? an interdisciplinary review of current issues in AI evaluation." In *Proceedings of the AAAI/ACM Conference on AI, Ethics, and Society*, vol. 8, no. 1, pp. 850-864. 2025 (and the many references therein). Even sophisticated and reflective takes on benchmarking tend to take for granted its "success" and continued pervasiveness (Moritz Hardt, *The Emerging Science of Machine Learning Benchmarks* https://mlbenchmarks.org/ ).

One admirable outlier, that recognises the need to move beyond benchmarks, is Ben Hutchinson, Negar Rostamzadeh, Christina Greer, Katherine Heller, and Vinodkumar Prabhakaran. "Evaluation gaps in machine learning practice." In *Proceedings of the 2022 ACM conference on fairness, accountability, and transparency*, 1859-1876, 2022.

[80] The power of benchmarks (that they scalarise everything) is their biggest weakness. One could argue that the problems of AI powered bureaucracy are nothing to do with "algorithms," and all to do with the objective function being optimised. In the same way that turning education into an exercise in passing tests spectacularly misses its primary value, likewise with algorithms.

[81] Confer Robert Williamson, "Process and Purpose not Thing and Technique," *Harvard Data Science Review* 2(3), 2020.

[82] Bruno Latour, *Pandora's hope: essays on the reality of science studies.* Harvard university press, 1999. For the history of the idea of a black box see P. von Hilgers, The History of the Black Box: The Clash of a Thing and its Concept, *Cultural Politics* 7(1), 41-58, 2001.

A common complaint about algorithms is that they are "black boxes"[83]. I wish they really were. The engineer's black box comes with data sheets and a manual that tells you how it will behave. But data can also be framed as a black box, and its manual would serve as its warrant.[84]

What might be done?
- Don't view data not as a fact, but as a process. *Show all your working!* Pass it on.[85]
- You would not put some arbitrary thing you found on the road in your mouth (unless you were 2 years old); why would you behave like a 2 year old with your LLM?
- *Provide some room for disagreement.* Make ML, like all good arguments, *controvertible.* In order to do this, stop taking so many "facts" for granted.

## AGENTS AND DELEGATION

Even hotter than ML is *agentic ML.*

But what does that mean?

Ultimately, it is about *delegation*. And in delegation, your delegate has *limited* autonomy, but you remain responsible.[86]

---

[83] The most common response to their perceived black box nature is to demand "explanations" (this is now popular enough to have become acronymised to as "XAI".) Or worse, to demand that one should "open the black box" (search the internet for this phrase to see its popularity). Amusingly, this spectacularly misses the entire point of black boxes: *good* black boxes are good precisely *because they do not need to be opened*; everything you need to know in order to use the box, is described in the associated documentation.

The problem with ML algorithms is not that they *are* black boxes; but that they are *bad* black boxes — nobody bothers to design them so that they can be documented, let alone puts much effort into effective documentation. I conjecture that this is simply a consequence of valorising performance on benchmarks above all else — you don't achieve "SOTA" on your favourite benchmark by writing a good manual. Contrast machine learning to mature technologies (say automobiles) where there are countless tradeoffs of narrow performance (say speed) against reliability and certainty of what the machine will do.

[84] While there is some work now on keeping track of data, and thus making models reversible, it is still a niche issue. See the literature on "data provenance" or "data journeys" and "data governance" (eg Sabina Leonelli and Niccolò (Editors), Data Journeys in the Sciences, Springer Nature 2020.) A specific example of ML engineers attempting to make the chains reversible is Mike Dreves, Gene Huang, Zhuo Peng, Neoklis Polyzotis, Evan Rosen, and Paul Suganthan G. C.. 2020. From Data to Models and Back. In *International Workshop on Data Management for End-to-End Machine Learning (DEEM'20)*, June 14, 2020. For more literature look at the now voluminous works citing T. Gebru, J. Morgenstern, B. Vecchione, J.W. Vaughan, H. Wallach, H. Duame, K. Crawford. Datasheets for datasets. *Communications of the ACM.* 19;64(12):86-92 November 2021.

[85] View data instead as "capta" (taken) — see Rob Kitchin, *The data revolution: Big data, open data, data infrastructures and their consequences.* Sage, 2014. Or view it as "sublata" (lifted up) — see Bruno Latour, Ibid.

[86] This simple notion cuts through a lot of nonsense talked about regarding "autonomous" ML systems. For details, see the discussion of, and pointers to, the literature on "second-order responsibility" in Robert C. Williamson, The AI of ethics, In *Machines we Trust: Perspectives on Dependable AI*, pages 139–160. MIT Press, 2021.

When users interact with software, they think they are having a conversation with the machine; but it is far more accurate to say they are having an indirect conversation with the (human) designer of the machine. (Alan Blackwell, *Moral Codes: Designing Alternatives to AI*, MIT Press 2024; pages 65ff)

The point is that ML systems are *always* commissioned by a *person*, and that person is thus responsible. This is nothing weird. After all, we delegate our *moral philosophy* to *lumps of concrete! ...* in the form of speed bumps in roads.[87]

## SUMMARY SO FAR

Machine learning is a style of reasoning, and is as rhetorical as any other. It
- Takes data as *fact* (not a core object of enquiry)
- Presumes the data is "*random*" (as an omnibus sanitisation protocol)
- Purports to learn *representations* of the world (from the "intrinsic structure of data")
- Presumes that *knowing* the world suffices to *control* it
- Takes *categories* as features of the *world* (to avoid grappling with the hard choice)
- Avoids grappling with the tension between the *individual* and the *aggregate*
- Confuses and conflates *data* and *information*
- *Valorises method above all*
- Judges methods solely via canned "*benchmarks*"
- Makes black boxes, without providing the associated data-sheets.
- Construes its products as fully autonomous, when it is mere partial delegation.

It has honed its style of reasoning so that the style is invisible. It has thus successfully turned itself into a *self-perpetuating thought-style* — in other words, a "discipline"[88].

## It ain't the way you do it, its what you do

Enough on method and technologies! Technologies are not *ends*. They are *means.*

In the last part of my talk I would like to make a few observations about ends.

There are plenty of ends to which ML has been put that are wonderful, and ML has unquestionably contributed to the world in positive ways. *But...*

---

[87] Bruno Latour, Ibid.

[88] It has converged to "an argumentation format that restricts (i) word usage, (ii) borrowings permitted from other disciplines, and (iii) appropriate contexts of justification/discovery (for example, some claims may be grounded on "reason alone," some on unaided perception, some on technically aided perception)."
It has also managed to "define and maintain the 'normal' state of objects in the domain. This involves experimental and textual techniques for foregrounding the problematic claims under study against a background of claims that are stipulated to be unproblematic." [Steve Fuller, *Social Epistemology,* (2nd edition), Indiana University Press, 2002, pages 191 and 197].

A careful history of ML as a discipline remains to be written.

## LARGE LANGUAGE MODELS

Front of mind, and motivating for this conference, are "large language models" which purport to capture knowledge about the world.

As leading ML scholar Tom Dietterich has observed[89], LLMs are not a *knowledge base*, but a *statistical model of a knowledge base.*

LLMs "generate" their outputs by sampling from their posterior distribution! An LLM is very complex model of its inputs. But it is a *probabilistic* model.

It is literally a *probability distribution*. The outputs it "generates" are the *most likely ones* under that model — a fancy version of autocomplete.

Yes, after the investment of countless billions of dollars and petajoules of energy, we have built a cliché generator[90] that is "trained to talk like the internet."[91]

LLMs offer next to nothing in terms of warrants for their arguments.

They are a library made from all the books found at a recycling plant, with pages removed from their bindings, all stirred together, and used without taking account of their provenance.

The apparent knowledge that they have is simply a statistical summary of the texts for which they have counted relative frequencies of different phrases (well actually tokens).

*But we can think of LLMs differently.*

---

[89] https://www.youtube.com/watch?v=e8vg1vin78U

[90] This is not so unfair: LLMs literally (but randomly) pick the "most likely" continuation of a prompt; what else is that other than a cliché? One might go further and co-opt the more pointed phrase "thought terminating cliché" as a more accurate description. We would be arguably better served valorising the *language* in "LLM" rather than *model*. As Marshall McLuhan put it "Language does for intelligence what the wheel does for the feet and the body." *The Gutenberg galaxy: The making of typographic man,* University of Toronto Press, 2011. Of course, McLuhan's slogan only works if you actively exercise your intelligence with language, rather than mindlessly delegating to a statistical algorithm…

There is an increasing literature now on LLMs as cliché generators [Ihor Rudko, and Aysan Bashirpour Bonab. ChatGPT is incredible (at being average). *Ethics and Information Technology* 27, no. 3, 2025] and bullshit generators [Michael Townsen Hicks, James Humphries, and Joe Slater. ChatGPT is bullshit. *Ethics and information technology* 26, no. 2, 1-10, 2024]

[91] The nice phrase is due to Alan Blackwell, *Moral Codes: Designing Alternatives to AI*, MIT Press 2024. He argues that the mistake is to take Alan Turing's imitation game as a business strategy and that "the companies 'winning' the test are able to do so only when they make their customers more stupid, while also needlessly consuming the precious resource of conscious human attention." (page 125)

The *original* language models were simply indices; pointers that helped a reader find what they were after. They thus provided an interface (API) to the data in the books.

They were implemented at scale, in specialised large "data centres".

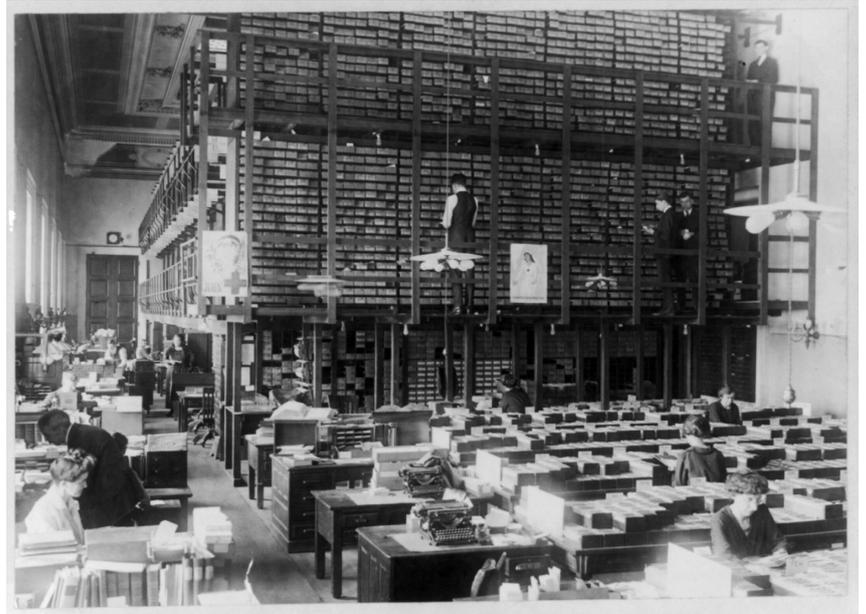

Perhaps we should try to build better artificial librarians, who do not presume to *answer* our questions, and certainly do not try to *persuade* or *manipulate* us, but rather, simply help us find information which we can then judge ourselves, and make up our *own* minds, thus *preserving our autonomy* … which brings me to my last point.

## AUTONOMY: PERSUASION VS MANIPULATION

Paying close attention to information flows allows us to draw an importance distinction that is strangely ignored in the discussion of the impact of AI systems.

Much is made of the *privacy* of data in ML systems. But the usual notions of personal information don't help, since private traits can be accurately predicted from unprotected data such as "likes"[92].

And even your personality[93] can be accurately predicted from your browsing history[94].

Karina Vold and Jess Whittlestone grappled with this, and made the distinction between:

*Persuasion* is when you *know* you are being pushed; manipulation when you are *not aware*.

*Persuasion* is manifest to the subject, and is credibly rejectable.

---

[92] Michal Kosinski, David Stillwell, Thore Graepel, Private traits and attributes are predictable from digital records of human behavior, *Proceedings of the National Academy of Sciences* 110 (15) 5802-5805, Apr 2013.

[93] The "big 5 dimensions" for example.

[94] M. Kosinski, P. Kohli, D. J. Stillwell, Y. Bachrach, & T Graepel. Personality and website choice. *ACM Web Science Conference*, Evanston, Illinois, USA, 251–254, 2012.

*Manipulation* is insidious precisely because it is *not* manifest, and thus not rejectable. ML can make statistical models to tune information sent to you. The more accurate the models thus produced, the more accurate and effective the targeting.[95] Vold and Whittlestone observed that such "personalised targeting is particularly likely to be manipulative."[96]

This matters, as it underpins some of the largest businesses built upon machine learning…

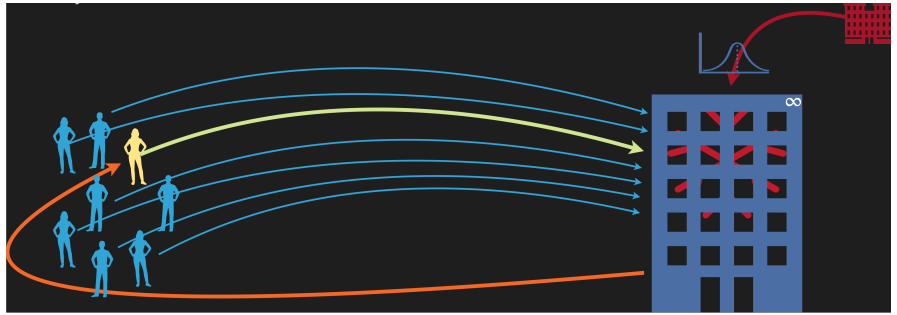

## MANIPULATION AS A SERVICE — AN ML-POWERED BUSINESS

Gerald Dworkin observed:

> One way of interfering with your autonomy is to deceive you. This interference with information is, however, just the opposite kind from that involved in interference in privacy. What is controlled is the information coming *to you,* not the information coming *from you.*[97]

How does this matter? Consider the scenario pictured above. Here's you (in yellow). You provide some information to a powerful corporation (green line)[98]. But so do lots of other people (blue lines). With this information the corporation builds a statistical model (dark blue distribution). They then sell an API to that model to a third-party (red line). Using this API, the third-party can send messages via the Corporation along the orange line that

---

[95] And "effective" here means "the more that behaviour is modified to the satisfaction of the advertiser."

[96] Karina Vold and Jess Whittlestone, Privacy, Autonomy, and Personalised targeting: Rethinking How Personal Data is Used, in *Report on Data, Privacy, and the Individual in the Digital Age*, by IE University's Center of the Governance of Change, 2019.

[97] Gerald Dworkin, *The theory and practice of autonomy.* Cambridge University Press, 1988, page 104.

[98] An aside: why anyone willingly subjects themselves to such "cognitive cigarette'' businesses baffles me. The name suits: they are considered "cool"; "everyone" uses them; they are highly addictive; vast profits are made by the corporations purveying them; they are extremely harmful (but subtly so — the harms accrue long after the use); and you feel much better when you quit your addiction. Perhaps these businesses will eventually be regulated as part of public (mental) health initiatives in the same way that smoking has (gradually) been so regulated (in the face of the 6-8 million deaths per year that it causes).

My speculation is that in general we believe that we can ingest *any* information into our minds without it doing things to us which we are unaware of. The irony is we take great care to not insert a found memory stick into our computer (because of the harm that the information on the stick might cause); but we happily ingest arbitrary information into our minds, which do not have any malware scanners, without worrying about the harm that it will cause. We are good at data hygiene for our computers; but we are very bad at information hygiene for own minds. We have technologies that prevent informational manipulation and abuse of *websites* (e.g. Cloudflare WAF); but we lack tools to help maintain information hygiene for our *own minds.*

statistically are likely to influence you. Thus you ("probably") change your mind/position relative to what you held previously. Note that *none* of your secret information has been released. Neither the blue nor the red company care who you are. All they need to know is that their intervention, that is the message that they sent you on this orange wire, will be effective, by which I mean it will change your mind and move you to a different position. That's the business model of manipulation-as-a-service.

In this scenario, you are not being *persuaded.* You are being *manipulated.*

It is subtle: the *modelling* is done at an *aggregate* level. The *impact* (effect) occurs at the *individual* level.

Neither of the companies needs to know who you are; your privacy is preserved.

They only care they have made another conversion (i.e. another manipulation).

They can then brag to their paying customers how well they have segmented the population, and thus the precision targeting they can offer to the customer's payload.[99]

## INFORMATION FLOWS UNDERPINNING MAAS

MAAS is not invincible. It can be disrupted by breaking the feedback loop at any point.

One can cut the green wire by not providing any information, or providing noisy information.[100]

One can cut the orange wire by not reading what they send you[101].

---

[99] MAAS is more commonly called by the euphemism "advertising". There are now good critical histories of it (e.g. Mark Bartholemew, *Adcreep: The Case Against Modern* Marketing, Stanford Law Books 2017; Chris Jay Hoofnagle, Ashkan Soltani, Nathaniel Good, and Dietrich J. Wambach. "Behavioral advertising: The offer you can't refuse." *Harvard Law and Policy Review*. 6, 273—296, 2012; , Alonso Villarán. "Irrational advertising and moral autonomy." *Journal of Business Ethics* 144, no. 3, 479-490, 2017.)

There is now also solid evidence of the aggregate harm that it does (Chloe Michel, Michelle Sovinsky, Eugenio Proto, and Andrew J. Oswald. "Advertising as a major source of human dissatisfaction: Cross-national evidence on one million Europeans." In *The economics of happiness: how the Easterlin paradox transformed our understanding of well-being and progress*, pp. 217-239. Cham: Springer International Publishing, 2019.)

[100] The browser plugin TrackMeNot does a simplistic version of this. Other mitigations are conceivable and already available in primitive forms (some web browsers already programmatically interfere with website fingerprinting methods, hiding IP and MAC addresses etc)

[101] Browser plugins that block advertisements, hide sponsored search ads, strip out tracking code from search results, control what javascript can run in your browser, or turn off AI search features all contribute to this.

One can disrupt the red wire by programmatically interfering with the economics (having your computer automatically click all the ads for you)[102].

In principle, one could regulate the orange wire, if one were prepared to accept that "social media companies" are "media companies."

## THERE IS ALWAYS A CHOICE

Much of the commentary on AI/ML is infected with "technological determinism"[103] — new technologies will continue to develop on a pre-determined trajectory, and there is nothing we can do but watch, whinge and wail…

Or, instead, we can choose to "dispute the indisputable."[104]

We can ask, with Alan Blackwell, "Why are we making software we don't need?"[105]

So let me finish with a speculation.

ML, like *all* technologies is "neither good nor bad, neither is it neutral."[106]

---

[102] The browser plugin AdNauseam does a fine job of this.

It is intriguing to speculate how this (and the previous mentioned browser enhancements) could be made substantially more powerful by the use of ML technology deployed in the service of the user, rather than being deployed in order to monetise the user (learning from societal scale use what the latest manipulation attempts are, and systematically blocking them). A corporation that managed to protect people at scale from these various "cognitive cigarettes" (see earlier footnote) could conceivably change the economics of the internet. Perhaps an innovative *bank* would do so. After all, banks look after *your* money, and nowadays money is largely *bits*. They are among our most trusted corporations; at least it seems reasonable to trust your bank more than an advertising company. Why not have the bank look after *all* your valuable bits (information) and protect *you* from harm (as they do, for example, by indemnifying you from much credit card fraud)?

[103] The phrase is widely used in studies of social-technical systems, as a quick internet search will confirm. A starting point is Thomas P. Hughes, The evolution of large technological systems. In *The social construction of technological systems.* Cambridge MA: MIT Press. pp. 51-82, 1987.

One could write a whole book on AI from this perspective. And indeed, one has been so written, which, *inter alia*, elegantly and compellingly refutes the determinist arguments: Emily Bender and Alex Hanna, *The AI Con: How to Fight Big Tech's Hype and Create the Future We Want,* Harper, 2025.

[104] The phrase comes from Alain Desrosières, *The Politics of Large Numbers: A History of Statistical Reasoning,* Harvard University Press, 1998, pp323ff. He uses it to signal a resistance to the taking of data as fact. But we can also apply it to the pushing back against the determinist narrative of ML technology that Bender and Hanna so ably critique.

[105] Page 3 of Alan Blackwell, *Moral Codes: Designing Alternatives to AI*, MIT Press 2024. Why not build systems that allow us to talk back? Not through the narrow gate of a "chat" interface, but through better languages (See chapter 10 of Blackwell, op. cit.). What not make technologies to facilitate better connections with others; systems that facilitate good conversations, rather than being persuaded and manipulated.

[106] This memorable line about technologies is known as "Kranzberg's law", and was stated by Melvin Kranzberg in: Technology and History: "Kranzberg's Laws", *Technology and Culture,* Vol. 27, No. 3, 544-560, 1986.

There *are* alternatives, which have been long envisaged. Ivan Illich's *Tools for Conviviality*[107] was a major motivation for the development of the personal computer[108].

We could take that as an alternate starting point for ML systems that interact with people.

But *how* can we do this? How can we lift the human spirit, and contribute to human flourishing, rather than along with "reverse Andrew Carnegies"[109] creating more "reverse centaurs"[110] — beings with the body of a human, but with their mind manipulated and controlled by AI? Who can help us!?

Alan Kay offered a compelling answer 50 years ago:

> Children who have not yet lost much of their sense of wonder and fun have helped us to find an ethic about computing: Do not automate the work you are engaged in. only the materials. If you like to draw, do not automate drawing; rather, program your personal

---

[107] Ivan Illich, *Tools for Conviviality*, Fontana 1975.

[108] Anonymous, *Lee Felsenstein and the Convivial Computer,* 23 July 2007, http://conviviality.ouvaton.org/spip.php?article39 ; Felsenstein's own words make it clear that he was inspired by Illich, and he believed "that computer hardware can also be designed and handled in a convivial fashion"; see Lee Felsenstein, *Tom Swift Lives!,* 1974; available at https://archive.org/details/felsenstein-tom-swift-lives . See also John Markoff, *Machines of Loving Grace: The Quest for Common ground Between Humans and Robots*, 2015 and his earlier *What the Dormouse Said: How the Sixties Counterculture Shaped the Personal Computer Industry*, Penguin 2005.

The early days of personal computing embraced the convivial perspective; but it is absent from too many internet-scale machine learning applications (there are, of course, plenty of admirable exceptions, such as ML-based language translators, but there are plenty of horrific examples of non-convivial technologies which solely exist to maximise profits). But I think this widespread absence (of convivial uses) is an accident of history, rather than anything inevitable.

One simple way to convey what I have in mind is to distinguish between AI (*Artificial Intelligence*) which seeks to automate everything, and IA (*Intelligence Augmentation*) which seeks to aid and augment human intelligence without replacing it.

Perhaps surprisingly (to a contemporary audience), this distinction was made in what is arguably one of the very first books on AI, W. Ross Ashby's *An Introduction to Cybernetics,* Chapman and Hall, 1957. It was a direct inspiration for the seminal work by Doug Englebart (inventor of the computer mouse, amongst other things): Douglas C. Engelbart, *Augmenting human intellect: A conceptual framework*, Technical Report SRI Project 3578 / AFOSR-3223, Stanford Research Institute, 1962. For a more recent analysis of "augmentation," see Peter Skagestad, Thinking with machines: Intelligence augmentation, evolutionary epistemology and semiotic. *Journal of Social and Evolutionary Systems* 16 (2), 157–180, 1993.

[109] Andrew Carnegie accumulated great wealth in an earlier industrial era, and then used much of it to fund a vast network of libraries, where he insisted the stacks be open so that people could freely access the (knowledge in) books. The current generation of industrial plutocrats use their wealth to appropriate the contents of libraries (avoiding any compensation if at all possible), locking up the contents, and then selling the resulting statistical summary in order to accumulate even further wealth. In so doing, they are effectively reversing the civic contribution Carnegie explicitly set out to make, so it seems not unreasonable to call them "reverse Carnegies".

[110] Creatures with human bodies, but without the human mind. That is, humans in the thrall of AI. The powerful metaphor is due to Cory Doctorow, *Enshittification: Why Everything Suddenly Got Worse and What to Do About It*, MCD Farrar, Straus and Giroux 2025.

computer to give you a new set of paints. If you like to play music, do not build a "player piano"; instead program yourself a new kind of instrument.[111]

We can extrapolate: if you like to converse, do not *automate*[112] conversation, but build ways of *connecting* with a richer set of people, and *facilitating* better human conversations[113].

This has little to do with the underlying technologies *themselves* (which are always *means*), but rather with the *ends* to which they are put, and the *way* they are used.

This obviously matters, especially for technologies of persuasion and manipulation…

Let me end my speculation, and my talk, with the empowering observation of Lewis Mumford which we would do well to remember:

> Nothing that man has created is outside his capacity to change, to remould, to supplant, or to destroy: his machines are no more sacred or substantial than the dreams in which they originated.[114]

---

[111] Page 244 of Alan Kay, Microelectronics and the Personal Computer, *Scientific American* 237(3), 230—245, September 1977. The wonderful thing about children, is that, if young enough, they are not infected by the disease of wealth-worship. Thus they can focus on living. Kay was presciently aware of the danger of the "human propensity to place faith in and assign higher powers to an agency that is not completely understood." The relevance to modern ML systems is too obvious to belabour. The history of the idea of "tools for thought" is explored by Howard Rheingold, *Tools for Thought: The History and Future of Mind-Expanding Technology,* MIT Press, 2000.

It is instructive to re-read the works of user interface designers from decades ago when the goal was for such convivial tools, looking at how they approached the problem. We could readily apply many of their lessons today. See for example: Stuart K. Card and Thomas P. Moran, *User Technology: From Pointing to Pondering, A history of personal workstations.* 489-526, 1988; Gerhard Fischer, Beyond "couch potatoes": From consumers to designers. In *Proceedings. 3rd Asia Pacific Computer Human Interaction*, 2-9. IEEE, 1998; Sherry Turkle and Seymour Papert. Epistemological pluralism: Styles and voices within the computer culture. *Signs: Journal of women in culture and society* 16, no. 1, 128-15, 1990; and Sherry Turkle, *The Second Self: Computers and the Human Spirit* (Twentieth Anniversary Edition), MIT Press 2005 (original edition published in 1984).

[112] It would be an interesting exercise for psychologists to understand *why* the engineers and scientists who build ML systems default to the presumption that the task at hand is to *automate* what was done by humans, in the name of "efficiency," rather than that of building tools and technologies that *augment* humans, enabling human flourishing. My speculation is that they have forgotten how to be children (per Alan Kay).

[113] Of course networked computing has offered many such opportunities, although some of them are now poisoned by the imposition of MAAS business models. *But this is not inevitable.* It is a good exercise (and not at all obvious) to ask how we can use ML technology for facilitating human conversation, to promote the ideals of conviviality and flourishing. As a challenging stretch goal, we might ask: what use could ML be put in societies such as that of the Kesh, imagined by Ursula Le Guin in her wonderful *Always Coming Home* (Harper and Row, 1985).

What matters is the degree of flexibility and autonomy that the new technologies leave to people. One way of thinking about that is in terms of the "workmanship of risk" (where there remain many things that can go wrong if you, the human, do not do things right), versus the "workmanship of certainty" which seems to be the default framing of all those who think that the single best thing to do with ML is "automate" every human task (without asking themselves what the humans will then do). The distinction between workmanship of risk versus certainty is from David Pye, *The Nature and Art of Workmanship,* Herbert Press 2007.

[114] Lewis Mumford, *The Condition of Man.* Martin Secker and Warburg, 1944.